\documentclass[letterpaper, 10 pt, conference]{ieeeconf}  

\IEEEoverridecommandlockouts                              

\usepackage{tabularx}
\usepackage{amsmath,amsfonts}

\usepackage{threeparttable}
\usepackage{cite}
\usepackage{multirow}
\usepackage{longtable}
\usepackage{graphicx}
\usepackage{caption}
\usepackage{subcaption}
\usepackage{float}

\usepackage{soul}
\usepackage{url}
\usepackage{booktabs}
\usepackage{algorithm}
\usepackage{algorithmic}
\usepackage[switch]{lineno}

\usepackage{lineno}
\usepackage[hidelinks]{hyperref}

\hypersetup
{
 unicode=false,     
 pdftoolbar=true,    
 pdfmenubar=true,    
 pdffitwindow=false,   
 pdfstartview={FitH},  
 pdftitle={My title},  
 pdfauthor={Author},   
 pdfsubject={Subject},  
 pdfcreator={Creator},  
 pdfproducer={Producer}, 
 pdfkeywords={keywords}, 
 pdfnewwindow=true,   
 colorlinks=true,    
 linkcolor=red,     
 citecolor=blue,    
 filecolor=magenta,   
 urlcolor=cyan      
}

\title{\LARGE \bf
Integrating Multi-Modal Sensors: A Review of Fusion Techniques for Intelligent Vehicles
}

\author{Chuheng~Wei$^{*}$, ~\IEEEmembership{Student~Member,~IEEE},
        Ziye Qin,~\IEEEmembership{Student~Member,~IEEE},
        Ziyan Zhang,\\ Guoyuan~Wu,~\IEEEmembership{Senior~Member,~IEEE},
        Matthew~J.~Barth,~\IEEEmembership{Fellow,~IEEE},
\thanks{Chuheng Wei, Ziyan Zhang, Guoyuan Wu, and Matthew J. Barth are with the College of Engineering, Center for Environmental Research and Technology, University of California at Riverside, Riverside, CA, 92507.}
\thanks{Ziye Qin is with the School of Transportation and Logistics, Southwest Jiaotong University, Chengdu, China.}
\thanks{$^{*}$Corresponding author. e-mail: chuheng.wei@email.ucr.edu}
}

\begin{document}

\maketitle
\begin{abstract}
Multi-sensor fusion plays a critical role in enhancing perception for autonomous driving, overcoming individual sensor limitations, and enabling comprehensive environmental understanding. This paper first formalizes multi-sensor fusion strategies into data-level, feature-level, and decision-level categories and then provides a systematic review of deep learning-based methods corresponding to each strategy. We present key multi-modal datasets and discuss their applicability in addressing real-world challenges, particularly in adverse weather conditions and complex urban environments. Additionally, we explore emerging trends, including the integration of Vision-Language Models (VLMs), Large Language Models (LLMs), and the role of sensor fusion in end-to-end autonomous driving, highlighting its potential to enhance system adaptability and robustness. Our work offers valuable insights into current methods and future directions for multi-sensor fusion in autonomous driving.
\end{abstract}

\section{Introduction}

\subsection{Background}
Perception serves as the cornerstone of autonomous driving systems, enabling vehicles to understand and interact with their surroundings safely and effectively~\cite{wang2019multi}. Sensors play a pivotal role in this perception process, as they collect critical environmental data necessary for tasks such as adaptive cruise control (ACC), lane keeping, and collision avoidance~\cite{fung2017sensor}. However, the limitations of individual sensors, including restricted range and susceptibility to environmental factors, necessitate cooperative sensing~\cite{kocic2018sensors}.

Relying solely on one type of sensor often results in performance bottlenecks due to inherent weaknesses. For example, cameras struggle under low-light conditions~\cite{wei2024feature}, LiDAR can be hampered by adverse weather~\cite{kutila2018automotive}, and Radar provides limited resolution for fine-grained object recognition~\cite{lu2010radar}. To overcome these challenges, multi-sensor fusion has emerged as a vital approach in autonomous driving. By integrating data from multiple sensor modalities, multi-sensor fusion provides a more comprehensive and robust understanding of the environment, enhancing the reliability and safety of autonomous systems~\cite{kocic2018sensors}.

As the complexity of driving scenarios increases, the role of multi-sensor fusion becomes even more significant. Modern autonomous driving systems demand a seamless integration of data from heterogeneous sensors to address challenges such as adverse weather conditions, dynamic road environments, and real-time decision-making~\cite{yeong2021sensor}. This has fueled extensive research and innovation in multi-sensor fusion techniques, making it a critical area of study in the advancement of autonomous driving.

\subsection{Sensors in Autonomous Driving}
\paragraph{Camera}
Cameras are among the most widely used sensors in autonomous driving due to their ability to capture high-resolution images and detailed visual information~\cite{wang2021research}. They excel in recognizing objects~\cite{wei2023enhanced}, traffic signs~\cite{de1997road}, and lane markings~\cite{andrade2018novel}, providing rich semantic data crucial for decision-making. However, cameras are sensitive to lighting conditions and can perform poorly in scenarios such as nighttime driving or glare from direct sunlight.

\paragraph{LiDAR}
LiDAR sensors utilize laser beams to measure distances and generate precise 3D maps of the environment~\cite{wang2021research}. Their high spatial resolution and ability to operate independently of ambient light make them invaluable for object detection and localization tasks~\cite{zywanowski2020comparison}. Despite these advantages, LiDAR systems are costly, generate large volumes of data, and can be adversely affected by heavy rain or fog~\cite{kutila2018automotive}.

\paragraph{Radar}
Radar sensors are known for their robustness in adverse weather conditions, such as rain, fog, and snow~\cite{liu2021robust}. They excel in measuring object velocity and detecting moving targets at long ranges, making them particularly suitable for highway scenarios~\cite{chen2002speed}. However, Radar's resolution is lower compared to cameras and LiDAR, limiting its ability to provide detailed object classification and environmental mapping~\cite{kocic2018sensors}.

\begin{table}[h]
\centering
\caption{Comparison of Camera, LiDAR, and Radar capabilities.}
\label{tab:sensor_comparison}
\begin{threeparttable}
\begin{tabular}{l|ccc}
\hline
\textbf{Characteristic} & \textbf{Camera} & \textbf{LiDAR} & \textbf{Radar} \\
\hline
\textbf{Cost} & Low & High & Medium \\
\textbf{Data Volume} & Medium & High & Low \\
\textbf{Field of View} & Wide & Ultra-Wide†, Narrow‡ & Narrow \\
\textbf{Color Resolution} & High & None & None \\
\textbf{Rain Performance} & Poor & Medium & High \\
\textbf{Fog Performance} & Poor & Poor & High \\
\textbf{Night Performance} & Medium & High & High \\
\hline
\end{tabular}
\begin{tablenotes}
\item[†] Refers to mechanical LiDAR.
\item[‡] Refers to solid-state LiDAR.
\end{tablenotes}
\end{threeparttable}
\vspace{-1em}
\end{table}

The complementary strengths and weaknesses of these sensors underscore the necessity of sensor fusion in autonomous driving. By combining data from cameras, LiDAR, and Radar, a more holistic and resilient perception system can be achieved. Table~\ref{tab:sensor_comparison} provides a comparative evaluation of these sensors based on various criteria.

\subsection{Contributions}
Previous works such as \cite{fung2017sensor, ignatious2022overview, yeong2021sensor, kocic2018sensors, tang2023comparative} have summarized various autonomous driving algorithms, but rarely provide a formalized mathematical analysis of sensor fusion techniques. Additionally, few studies comprehensively discuss emerging hot topics in sensor fusion applications.

This paper makes the following contributions:
\begin{itemize}
\item \textbf{Formalized Sensor Fusion:} Provided a formalized mathematical summary of multi-sensor fusion in autonomous driving.
\item \textbf{Comprehensive Algorithm Review:} Reviewed state-of-the-art multi-sensor fusion algorithms, categorizing them into data, feature, and decision-level fusion while analyzing their methodologies, strengths, and limitations in autonomous driving.
\item \textbf{Datasets:} Conducted a thorough review of autonomous driving datasets, highlighting their suitability for multi-sensor fusion studies.
\item \textbf{Emerging Trends:} Discussed multi-sensor fusion in the context of emerging trends such as end-to-end autonomous driving and the integration of Vision-Language Models (VLM) and Large Language Models (LLM), providing insights into future directions for research.
\end{itemize}

\begin{figure}[ht]
    \centering
    \includegraphics[width=0.8\linewidth]{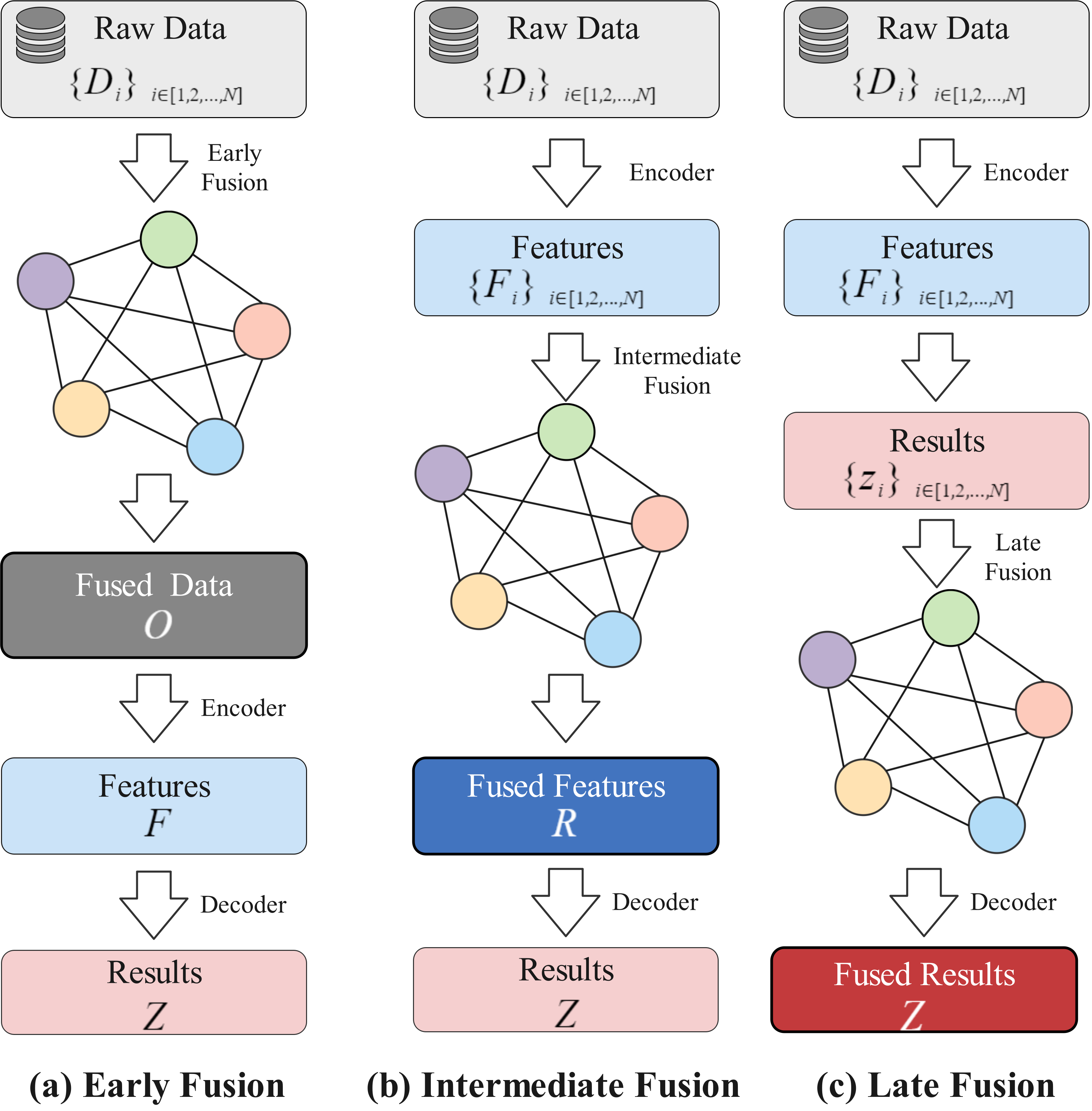}
    \caption{Architecture for Different Sensor Fusion Strategies}
    \label{fig:sensorfusion}
    \vspace{-1em}
\end{figure}

\section{Formalization}

Multi-sensor fusion in autonomous driving can be framed as a process that transforms raw data from multiple sensors into a unified output. Let \( D = \{D_1, D_2, \dots, D_n\} \) denote the set of raw data collected from \( n \) sensors, and \( Z \) represent the final fused output. The fusion process is formulated as:
\begingroup
\setlength{\abovedisplayskip}{2pt} 
\setlength{\belowdisplayskip}{2pt}
\begin{equation}
    Z = \Omega(D; \theta),
\end{equation}
where \(\Omega(\cdot; \theta)\) is the overall fusion function parameterized by \(\theta\), responsible for integrating sensor data for tasks such as object detection, tracking, or decision-making.

To maintain consistency across fusion strategies, we define \(\alpha\), \(\psi\), and \(\phi\) as parameter vectors associated with the fusion, encoding, and decoding functions, respectively. These parameters may vary in form depending on the fusion level. While not all parameters are explicitly instantiated in every equation, they represent tunable components within the respective functions.

The following subsections illustrate three common fusion strategies: Data-Level (Early), Feature-Level (Intermediate), and Decision-Level (Late). Figure~\ref{fig:sensorfusion} provides a visual representation of these strategies.

\subsection{Data-Level Fusion (Early)}
In data-level fusion, all raw sensor data are fused first, then transformed into features, and finally decoded:

\paragraph{Fuse Raw Data}

\begin{equation}
    O = G(D_1, D_2, \dots, D_m;\,\alpha),
\end{equation}
where \(G(\cdot;\,\alpha)\) merges the raw inputs into an intermediate representation \(O\).

\paragraph{Extract Features}
\begin{equation}
    F = E(O;\,\psi),
\end{equation}
where \(E(\cdot;\,\psi)\) encodes \(O\) into feature space.

\paragraph{Produce Output}
\begin{equation}
    Z = H(F;\,\phi),
\end{equation}
where \(H(\cdot;\,\phi)\) decodes the features \(F\) into the final output \(Z\).

\subsection{Feature-Level Fusion (Intermediate)}
Feature-level fusion separately encodes each sensor’s data, then fuses the resulting features before producing the final output:

\paragraph{Encode Each Sensor}
\begin{equation}
    F_i = E(D_i;\,\psi),
\end{equation}
where each \(F_i\) is a feature vector extracted from sensor \(D_i\).

\paragraph{Fuse Features}
\begin{equation}
    R = G(F_1, F_2, \dots, F_m;\,\alpha),
\end{equation}
where \(G(\cdot;\,\alpha)\) aggregates feature vectors \(F_1 \dots F_m\) into a fused representation \(R\).

\paragraph{Produce Output}
\begin{equation}
    Z = H(R;\,\phi),
\end{equation}
where \(H(\cdot;\,\phi)\) converts the fused features \(R\) into the final output \(Z\).

\subsection{Decision-Level Fusion (Late)}
Decision-level fusion first decodes each sensor’s feature into an intermediate prediction, then fuses these predictions to obtain the final result:

\paragraph{Encode Sensor Data}
\begin{equation}
    F_i = E(D_i;\,\psi),
\end{equation}
where \(E(\cdot;\,\psi)\) extracts features from the raw data \(D_i\).

\paragraph{Sensor-Specific Output}
\begin{equation}
    z_i = H(F_i;\,\phi),
\end{equation}
where \(H(\cdot;\,\phi)\) decodes each feature vector \(F_i\) into a sensor-specific output \(z_i\).

\paragraph{Final Fusion}
\begin{equation}
    Z = G(z_1, z_2, \dots, z_m;\,\alpha),
\end{equation}
where \(G(\cdot;\,\alpha)\) merges these sensor-specific outputs into the final fused result \(Z\).

\section{Fusion Methods for Multi-Sensor Integration}

Multi-sensor fusion is categorized by the integration stage within the perception pipeline into data-level, feature-level, and decision-level fusion, corresponding to early, intermediate, and late fusion techniques with different trade-offs in information preservation, computation, and robustness. Mix fusion methods further enhance adaptability by dynamically selecting fusion levels based on context. Table~\ref{table:DL-Fusion} summarizes fusion strategies, sensor combinations, and key applications.

\subsection{Data-Level Fusion (Early Fusion)}

Data-level fusion, or early fusion, combines raw sensor data at the beginning of the processing pipeline, enabling the model to learn from aligned sensor inputs directly, as illustrated in Figure~\ref{fig:sensorfusion}-(a) This approach is common in radar and camera fusion, where radar point clouds are projected onto the camera frame using a calibration matrix, creating a unified front-view perspective.
Since radar points often lie on the ground plane, further data augmentation and correction are applied to enhance accuracy. For instance, YOdar~\cite{kowol2021yodar} and CRF-Net~\cite{nobis2019deep} introduce a learnable vertical axis for each radar point, allowing the network to adaptively find the effective vertical depth coverage of radar points. Nabati et al. proposed RRPN~\cite{nabati2019rrpn}, a radar-based real-time region proposal network, which generates object proposals using radar data and predefined bounding boxes, replacing the RPN module in two-stage image detectors to accelerate processing.

As LiDAR sensors become more prevalent in autonomous vehicles, early fusion methods for camera-LiDAR data have advanced significantly. 
Qi et al. proposed Frustum PointNet~\cite{Qi_2018_CVPR}, which first detects objects in 2D images, converts them into 3D frustums via camera-LiDAR projections, and applies deep learning for 3D object detection. Wang et al. proposed PointAugmenting~\cite{wang2021pointaugmenting}, which enriches LiDAR point clouds with semantic features extracted from 2D image models and applies cross-modal data augmentation to improve detection performance under challenging conditions. 
A related method, PointPainting~\cite{Vora2020pointpainting}, enhances detection by projecting pixel-level semantic scores from an image segmentation network onto LiDAR points. Although PointPainting is often classified as a data-level or early fusion method due to the integration of semantic information before feature encoding, it is also considered by some to lie between data-level and feature-level fusion, as the semantic scores represent features extracted from multi-view images.

More complex early fusion methods incorporate multiple sensor modalities, such as camera, LiDAR, and radar. MVX-Net\cite{sindagi2019mvx} integrates RGB images and LiDAR data using simple yet effective fusion techniques like PointFusion and VoxelFusion within the VoxelNet architecture. Similarly, Lin et al.~\cite{lin2020depth} developed a dedicated stage to preprocess radar depth data before fusing it with image data, achieving improved depth estimation by leveraging complementary information from both radar and camera.

\begin{table*}[ht]
\caption{Overview of multi-sensor fusion methods and their applications}
\centering
\begin{threeparttable} 
\begin{tabular}{ccccccccc}
\toprule
\multirow{2}{*}{Method}&\multirow{2}{*}{Proposed Year}&\multicolumn{3}{c}{Fusion Level} &\multicolumn{3}{c}{Fused Sensors} & \multirow{2}{*}{Task}\\ 
\cmidrule(lr){3-5} \cmidrule(lr){6-8} 
 & & Data & Feature &  Decision & Camera &LiDAR&Radar &  \\ 
\midrule
Frustum PointNet~\cite{qi2018frustum} & 2018 & $\checkmark$     &   & 
&   $\checkmark$& $\checkmark$  &  &Object Detection \\

RRPN~\cite{nabati2019rrpn} & 2019 &   $\checkmark$  &   &   
&   $\checkmark$&   &$\checkmark$ &Object Detection \\

CRF-Net~\cite{nobis2019deep} & 2019 &   $\checkmark$  &   &     
&   $\checkmark$&   &$\checkmark$ &Object Detection \\

MVX-Net~\cite{sindagi2019mvx} & 2019 &  $\checkmark$   &   &     
&   $\checkmark$& $\checkmark$  & &Object Detection \\

~\cite{lin2020depth} & 2020 &   $\checkmark$  &   &    
&   $\checkmark$&   &$\checkmark$ &Depth Estimation \\

PointPainting~\cite{Vora2020pointpainting} & 2020 &  $\checkmark*$   &   &     
&   $\checkmark$& $\checkmark$  & &Object Detection \\

YOdar~\cite{kowol2021yodar} & 2021 &   $\checkmark$  &   &     
&   $\checkmark$&   &$\checkmark$ &Object Detection \\

PointAugmenting~\cite{wang2021pointaugmenting} & 2021 &  $\checkmark$   &   &     
&   $\checkmark$& $\checkmark$  & &Object Detection \\

MV3D~\cite{chen2017multi} & 2017 &     &  $\checkmark$ &     
&   $\checkmark$& $\checkmark$  &  &Object Detection \\


ContFuse~\cite{liang2018deep} & 2018 &     &  $\checkmark$ &     
&   $\checkmark$&$\checkmark$   &  &Object Detection \\

AVOD~\cite{ku2018joint} & 2018 &     &  $\checkmark$ &     
&   $\checkmark$&  $\checkmark$ &  &Object Detection \\

SO-Net~\cite{john2020so} & 2019 &     &  $\checkmark$ &     
&   $\checkmark$&   &$\checkmark$ &Semantic Segmentation \& Object Detection \\

MMF~\cite{liang2019multi} & 2019 &     &  $\checkmark$ &     
&   $\checkmark$& $\checkmark$  &  &Object Detection \\

~\cite{meyer2019sensor} & 2019 &     &  $\checkmark$ &     
&   $\checkmark$& $\checkmark$  &  &Semantic Segmentation \& Object Detection \\

3D-CVF~\cite{yoo20203d} & 2020 &     &  $\checkmark$ &     
&   $\checkmark$&  $\checkmark$ &  &Object Detection \\

EPNet~\cite{huang2020epnet} & 2020 &     &  $\checkmark$ &     
&   $\checkmark$&$\checkmark$   &  &Object Detection \\

PMF~\cite{zhuang2021perception} & 2021 &     &  $\checkmark$ &     
&   $\checkmark$& $\checkmark$  & &Semantic Segmentation \& Object Detection\\

VPFNet~\cite{zhu2022vpfnet} & 2022 &     &  $\checkmark$ &     
&   $\checkmark$&$\checkmark$   &  &Object Detection \\

Graph RCNN~\cite{yang2022graph} & 2022 &     &  $\checkmark$ &     
&   $\checkmark$& $\checkmark$  &  &Object Detection \\

SFD~\cite{wu2022sparse} & 2022 &     &  $\checkmark$ &     
&   $\checkmark$&$\checkmark$   &  &Object Detection \\

DeepFusion~\cite{li2022deepfusion} & 2022 &     &  $\checkmark$ &     
&   $\checkmark$& $\checkmark$  &  &Object Detection \\

AutoAlign~\cite{chen2022autoalign} & 2022 &     &  $\checkmark$ &     
&   $\checkmark$& $\checkmark$  &  &Object Detection \\

BEVfusion~\cite{liu2023bevfusion} & 2023 &     &  $\checkmark$ &     
&   $\checkmark$& $\checkmark$  &  &Object Detection \\

RCBEV~\cite{zhou2023bridging} & 2023 &     &  $\checkmark$ &     
&   $\checkmark$&   & $\checkmark$ &Object Detection \\

CRKD~\cite{zhao2024crkd} & 2024 &     &  $\checkmark$ &     
&   $\checkmark$&   & $\checkmark$ &Object Detection \\

GATR~\cite{luo2024gatr} & 2024 &     &  $\checkmark$ &     
&   $\checkmark$& $\checkmark$  &  &Object Detection \\

GAFusion~\cite{li2024gafusion} & 2024 &     &  $\checkmark$ &     
&   $\checkmark$& $\checkmark$  &  &Object Detection \\

HyDRa~\cite{wolters2024unleashing} & 2024 &     &  $\checkmark$ &     
&   $\checkmark$&   & $\checkmark$ &3D Semantic Occupancy Prediction \\

CO-Occ~\cite{pan2024co} & 2024 &     &  $\checkmark$ &     
&   $\checkmark$& $\checkmark$  &  &3D Semantic Occupancy Prediction \\

FOP-MOC~\cite{chavez2015multiple} & 2015 &     &   &  $\checkmark$   
&   $\checkmark$& $\checkmark$  &$\checkmark$  &Object Detection \\

~\cite{kunz2015autonomous} & 2015 &     &   &  $\checkmark$   
&   $\checkmark$& $\checkmark$  &$\checkmark$  &Object Detection \\

~\cite{du2018general} & 2018 &     &   &  $\checkmark$   
&   $\checkmark$& $\checkmark$  &  &Object Detection \\

~\cite{asvadi2018multimodal} & 2018 &     &   &  $\checkmark$   
&   $\checkmark$& $\checkmark$  &  &Object Detection \\

RoarNet~\cite{shin2019roarnet} & 2019 &     &   &  $\checkmark$   
&   $\checkmark$& $\checkmark$  &  &Object Detection \\

PI-RCNN~\cite{xie2020pi} & 2020 &     &   &  $\checkmark$   
&   $\checkmark$& $\checkmark$  &  &Object Detection \\

~\cite{kurapati2020multiple} & 2020 &     &   &  $\checkmark$   
&   & $\checkmark$  & $\checkmark$ &Multi Object Tracking \\

~\cite{melotti2020multimodal} & 2020 &     &   &  $\checkmark$   
& $\checkmark$  & $\checkmark$  &  & Object Detection \\

CLOCs~\cite{pang2020clocs} & 2020 &     &   &     $\checkmark$&
 $\checkmark$  & $\checkmark$  &  &Object Detection \\

Paigwar~\cite{paigwar2021frustum} & 2021 &     &   &  $\checkmark$   
&   $\checkmark$& $\checkmark$  &  &Object Detection \\

SCF~\cite{yu2024rethinking} & 2024 &     &   &    $\checkmark$ &
 $\checkmark$  & $\checkmark$  &  &Object Detection \\

~\cite{schlosser2016fusing} & 2016 & $\checkmark$     & $\checkmark$   &  $\checkmark$   
& $\checkmark$   & $\checkmark$   &  &Object Detection \\

~\cite{caltagirone2019lidar} & 2019 & $\checkmark$     & $\checkmark$   &  $\checkmark$   
& $\checkmark$   & $\checkmark$   &  &Object Detection \\

CenterFusion~\cite{nabati2021centerfusion} & 2021 &  $\checkmark$    &   & $\checkmark$    
& $\checkmark$   &   & $\checkmark$  &Object Detection \\

CRAFT~\cite{kim2023craft} & 2023 & $\checkmark$     &   &  $\checkmark$   
&  $\checkmark$  &   & $\checkmark$  &Object Detection \\

RCM-Fusion~\cite{kim2024rcm} & 2024 &     &  $\checkmark$  & $\checkmark$   
& $\checkmark$   & $\checkmark$   &  &Object Detection \\
\bottomrule
\vspace{-1em}
\end{tabular}
\begin{tablenotes}
    \item * Although often seen as early fusion, PointPainting\cite{Vora2020pointpainting} also shares traits with feature-level fusion due to the use of extracted semantic features.
\vspace{-1em}
\end{tablenotes}
\end{threeparttable}
\label{table:DL-Fusion}
\vspace{-1em}
\end{table*}

\subsection{Feature-Level Fusion (Intermediate Fusion)}

Feature-level fusion, as shown in Figure~\ref{fig:sensorfusion}-(b), combines sensor data at the feature extraction stage, effectively integrating complementary information from different modalities. This approach allows for richer feature representations and is currently the most widely used method in multi-sensor fusion, especially for applications in autonomous driving.

Initial works primarily focused on LiDAR-camera fusion for object detection. Chen et al. introduced MV3D~\cite{chen2017multi}, which generates 3D region proposals from LiDAR Bird's-eye view (BEV) data and fuses them with RGB image features to improve detection accuracy. Ku et al. proposed AVOD~\cite{ku2018joint}, a two-stage fusion approach that enhances bounding box regression and classification by integrating LiDAR and image features at the region proposal stage.

Subsequent approaches further enriched feature representation and adaptability. Liang et al. developed ContFuse~\cite{liang2018deep}, which integrates multi-scale LiDAR reflections while preserving spatial details. Huang et al. introduced EPNet~\cite{huang2020epnet}, aligning point features with semantic information via point-guided image fusion and consistency-enforcing losses. Yang et al. proposed Graph-RCNN~\cite{yang2022graph}, leveraging neighborhood graphs within region proposals for iterative message passing between image and point cloud features. Zhu et al. presented VPFNet~\cite{zhu2022vpfnet}, introducing "virtual" intermediate points to bridge the resolution gap between sparse point clouds and dense images. Chen et al. introduced AutoAlign~\cite{chen2022autoalign}, a learnable strategy that fuses image and voxel features through cross-attention for improved multi-modal consistency.

BEV-based fusion has become essential for 3D object detection. Liang et al. proposed MMF~\cite{liang2019multi}, integrating image features into BEV for high-accuracy detection and depth completion. Yoo et al. developed 3D-CVF~\cite{yoo20203d}, which projects 2D features into a smooth BEV space to refine proposals in LiDAR-camera fusion. Liu et al. introduced BEVFusion~\cite{liu2023bevfusion}, unifying multi-modal features in BEV to preserve geometric and semantic details efficiently. Zhao et al. extended this with CRKD~\cite{zhao2024crkd}, transferring feature knowledge from a LiDAR-camera model to a camera-radar model via knowledge distillation. Zhou et al. proposed RCBEV~\cite{zhou2023bridging}, addressing radar’s sparse, noisy points through a spatio-temporal encoder. Multi-modal fusion has also been applied in adverse conditions like fog. Zhang et al. introduced Transfusion~\cite{zhang2024transfusion}, a two-stage model that processes LiDAR and radar data independently before fusion, improving perception in low-visibility scenarios.

Beyond object detection, feature-level fusion supports semantic segmentation and 3D semantic occupancy prediction. John et al. proposed SO-Net~\cite{john2020so}, a radar-vision framework facilitating object detection and segmentation. Meyer et al. integrated camera and LiDAR features within a localized view for joint feature extraction. Zhuang et al. introduced PMF~\cite{zhuang2021perception}, dynamically weighting LiDAR depth and camera semantics for robust 3D segmentation. Extending feature fusion to 3D occupancy, Wolters et al. proposed HyDRa~\cite{wolters2024unleashing}, refining radar-camera fusion with radar-weighted depth consistency and a Height Association Transformer. Pan et al. introduced Co-Occ~\cite{pan2024co}, enhancing LiDAR-camera fusion via a geometry-aware module with KNN-based feature merging and volume rendering for improved 3D occupancy prediction.

\subsection{Decision-Level Fusion (Late Fusion)}

Decision-level fusion integrates sensor outputs at a higher abstraction level, where individual sensors provide their independent predictions, and the system combines them to reduce uncertainty and enhance robustness. This approach is particularly beneficial in autonomous driving, where fusing multiple sensor decisions improves overall perception and adaptability in dynamic environments.

Decision-level fusion remains widely explored in sensor-based perception, with most studies focusing on LiDAR-camera fusion. Asvadi et al.~\cite{asvadi2018multimodal} processed LiDAR point clouds into depth, reflectivity, and color maps, which were input into a YOLO-based network, followed by a scoring function and non-maximum suppression for refinement. Similarly, Du et al.~\cite{du2018general} introduced a model-based approach, refining 3D structures using predefined car models and a 2D CNN for enhanced accuracy. Shin et al.~\cite{shin2019roarnet} developed RoarNet, a two-stage pipeline estimating object poses from monocular images before 3D refinement. To improve pillar-based detection, Paigwar et al.~\cite{paigwar2021frustum} combined visual and point cloud features, enhancing localization in sparse data. Meanwhile, Melotti et al.~\cite{melotti2020multimodal} leveraged distance-based weighting to account for sensor performance variations. Addressing fusion consistency, Pang et al.~\cite{pang2020clocs} proposed CLOCs, aligning 2D and 3D detections through geometric and semantic constraints to improve efficiency. Further refining detections, Yu et al.~\cite{yu2024rethinking} introduced the Semantic Consistency Filter (SCF), which filters false positives based on 2D segmentation masks.

Beyond camera-LiDAR fusion, several studies have incorporated radar to enhance perception. Kunz et al.~\cite{kunz2015autonomous} introduced a hierarchical modular framework that probabilistically integrates radar, camera, and LiDAR data. Chavez-Garcia and Aycard’s FOP-MOC model~\cite{chavez2015multiple} applied an evidential fusion framework to combine classification outputs, improving object detection and tracking accuracy. Kurapati et al.~\cite{kurapati2020multiple} fused radar and camera outputs for multi-target tracking, using the Hungarian algorithm to refine associations and improve tracking stability.

By integrating sensor outputs at the decision level, these methods effectively enhance detection accuracy, mitigate individual sensor limitations, and improve robustness in real-world driving scenarios.

\subsection{Mix Fusion}

Mix fusion methods integrate data, features, and decisions from multiple sensors at various stages of the processing pipeline, offering a flexible and adaptive approach to multi-sensor perception in autonomous systems. Unlike strictly defined fusion strategies, mix fusion dynamically selects fusion levels based on task requirements and environmental conditions, allowing models to leverage the complementary strengths of different sensor modalities more effectively.

Schlosser et al.~\cite{schlosser2016fusing} were among the first to explore fusion of LiDAR and camera data within convolutional neural networks, demonstrating that different fusion levels impact performance uniquely. Their findings highlighted that combining multiple strategies enhances adaptability in complex scenarios. Expanding on this, Nabati et al. introduced CenterFusion~\cite{nabati2021centerfusion}, which integrates radar-based spatial features into an object detection pipeline by dynamically incorporating radar information at various processing stages rather than committing to a fixed fusion strategy. Similarly, Kim et al.~\cite{kim2023craft} proposed CRAFT, which employs a soft polar association and a spatio-contextual fusion transformer, enriching spatial and contextual information while making fusion decisions based on confidence levels.

Caltagirone et al.~\cite{caltagirone2019lidar} applied mix fusion in road detection, using an FCN-based approach that projects LiDAR point clouds onto the image plane to produce dense 2D road images while incorporating semantic segmentation outputs to refine road boundaries. More recently, Kim et al. introduced RCM-Fusion~\cite{kim2024rcm}, a radar-camera 3D object detection framework utilizing both feature-level and instance-level fusion. This method employs a Radar Guided BEV Encoder to convert camera data into a BEV format while refining object positions through radar-adaptive sampling, adjusting the fusion process based on sensor reliability.

In summary, mix fusion enables robust perception by adapting sensor integration across multiple levels, from raw data processing to final decision-making. By leveraging the strengths of various fusion techniques dynamically, mix fusion enhances detection accuracy and robustness in complex, dynamic environments, making it a promising direction for future multi-sensor fusion research.

\begin{table*}[!ht]
\centering
\setlength{\tabcolsep}{4pt} 
\caption{A detailed comparison of autonomous driving datasets for multi-sensor fusion.}
\begin{threeparttable} 
\begin{tabular}{lcccccc}
\hline
\textbf{Dataset} & \textbf{Year$^\dag$}  & \textbf{Camera} & \textbf{LiDAR} & \textbf{Radar}  & \textbf{Region/Platform} &\textbf{Tasks}  \\
\hline
KITTI~\cite{geiger2013vision} & 2013 & 4 & 1 & 0  & Germany & Object Detection, Object Tracking, Semantic Segmentation\\


ApolloScape~\cite{huang2018apolloscape} & 2018 & 6 & 2 & 0 & China & Object Detection, Object Tracking, Semantic Segmentation \\ 

Argoverse~\cite{chang2019argoverse} & 2019 & 7 & 2 & 0 & USA & Object Tracking, Motion Forecasting \\

Lyft Level 5~\cite{lyft2019} & 2019 & 7 & 3 & 5  & USA & Object Detection, Semantic Segmentation, Motion Prediction \\

Waymo Open Dataset~\cite{sun2020scalability} & 2019 & 5 & 5 & 0 & USA & Object Detection, Object Tracking \\

H3D~\cite{patil2019h3d} & 2019 & 3 & 1 & 0 & USA & Object Detection, Object Tracking \\

nuScenes~\cite{caesar2020nuscenes} & 2020 & 6 & 1 & 5 & USA, Singapore & Object Detection, Object Tracking \\

A*3D~\cite{pham20203d} & 2020 & 2 & 1 & 0 & Singapore & Object Detection \\ 

A2D2~\cite{geyer2020a2d2} & 2020 & 6 & 5 & 0 &  Germany & Object Detection, Semantic Segmentation \\ 

PandaSet~\cite{xiao2021pandaset} & 2020 & 6 &2 & 0& USA &  Object Detection \\


PixSet~\cite{deziel2021pixset} & 2021 & 3 & 1 & 1 & Canada & Object Detection \\

View-of-Delft~\cite{apalffy2022} & 2022 & 1& 1 & 1 & The Netherlands & Object Detection, Trajectory Prediction \\

DAIR-V2X-V~\cite{dair-v2x} & 2022 &  3 & 1 & 1 & China & Object Detection \\

TUMTraf V2X-V~\cite{zimmer2024tumtrafv2x} & 2024 & 1 & 1 & 0 & Germany & Object Detection, Semantic Segmentation \\

V2XReal-VC~\cite{xiang2024v2x} & 2024 & 4 & 1 & 0 & USA & Object Detection \\
\hline
\end{tabular}
\begin{tablenotes}
    \item $^\dag$ The publication year refers to the year the dataset was introduced in the respective paper.
\vspace{-1em}
\end{tablenotes}
\end{threeparttable} 
\label{dataset}
\vspace{-1em}
\end{table*}

\section{Datasets}
In this section, we introduce several available datasets collected from multi-sensor setups deployed on vehicles. The discussion focuses primarily on the use of cameras, radars, and LiDARs in each dataset and their potential applications to autonomous driving tasks. To obtain a valid dataset, several key steps are required: sensor deployment, sensor calibration, data collection, time synchronization, and data annotation. As summarized in Table \ref{dataset}, these datasets are sourced from vehicles operating across North America, Europe, and Asia, facilitating research into autonomous driving within diverse traffic scenarios. All of the vehicles used for data collection are equipped with at least cameras and LiDARs; however, radars are less commonly included, which limits the capability of some datasets to accurately sense distance and speed. Consequently, many datasets are less suitable for motion forecasting tasks. Each dataset has unique characteristics that make it valuable for specific research applications. For example, KITTI~\cite{geiger2013vision}, the first large-scale open-source dataset, remains widely used and serves as a benchmark for the format of related datasets. KITTI covers a wide range of scenarios, including highways and urban roads. Datasets like Argoverse~\cite{chang2019argoverse} and Lyft Level 5~\cite{lyft2019}, equipped with multiple cameras, offer 360-degree image data, enabling comprehensive environmental perception. Notably, the Argoverse dataset is the first large-scale dataset to include a semantic vector map, enhancing its utility for 3D tracking and motion forecasting. Meanwhile, the Lyft Level 5 dataset provides 1,118 hours of recorded data, making it the largest dataset for motion prediction and a rich resource for exploring diverse traffic scenarios. The A2D2 dataset~\cite{geyer2020a2d2} offers additional CAN bus information, which can be leveraged for end-to-end autonomous driving development. Furthermore, recent cooperative perception datasets, such as DAIR-V2X-V~\cite{dair-v2x}, TUMTraf~\cite{zimmer2024tumtrafv2x} , and V2XReal-VCC~\cite{xiang2024v2x}, integrate data from vehicle-mounted sensors, enabling advanced research into multi-sensor fusion.

\section{Discussion and Future Directions}

\subsection{Algorithmic Perspectives}

Feature-level fusion, the most widely used approach in multi-sensor fusion, primarily follows “Shallow Feature Intermediate Fusion”, where sensor data is aligned based on spatial correspondences, merely merging positional information without fundamentally transforming features. While effective, this approach limits cross-modal interactions, as features largely retain their original characteristics with minimal modifications beyond alignment and concatenation. Future research should explore “Deep Feature Intermediate Fusion”, where feature representations undergo learned transformations, such as convolutional operations, cross-modal attention, or adaptive re-weighting. This deeper integration can enhance feature synergy across modalities, improving robustness and adaptability in complex driving environments.

Advancements in hardware and algorithms enable the exploration of other fusion strategies. Data-level fusion is becoming more feasible with efficient preprocessing and high-throughput data handling. Similarly, decision-level fusion benefits from improved sensor-specific algorithms and contextual integration across modalities, enhancing perception robustness. Future research should move beyond structural integration, leveraging feature transformations for richer cross-modal interactions and deeper semantic fusion.

\subsection{Dataset Considerations}
The availability and diversity of datasets are critical for advancing multi-sensor fusion research. Current datasets often focus on specific sensor combinations, such as camera-LiDAR or camera-Radar, and may lack comprehensive multi-modal coverage for real-world scenarios. There is an increasing need for datasets that capture challenging conditions, including adverse weather, nighttime environments, and complex urban settings. Furthermore, datasets that provide synchronized, high-quality data from multiple modalities can enable more effective training and evaluation of fusion algorithms, accelerating progress in the field.

\subsection{Sensor Fusion in End-to-End Autonomous Driving}
In the context of end-to-end autonomous driving systems, multi-sensor fusion algorithms are expected to evolve towards seamless integration within neural architectures. The goal is to enable direct mapping from sensor inputs to driving actions while preserving interpretability and reliability. End-to-end fusion frameworks must address challenges such as handling diverse sensor data representations, ensuring real-time performance, and maintaining robustness under dynamic and unpredictable conditions. As these systems mature, they are likely to incorporate hybrid fusion strategies, leveraging the strengths of each fusion level to optimize decision-making.

\subsection{Sensor Fusion in VLM and LLM Integration}
The emergence of Vision-Language Models (VLMs) and Large Language Models (LLMs) offers exciting opportunities for advancing multi-sensor fusion. These models can facilitate cross-modal understanding by incorporating semantic context into perception tasks, enabling autonomous systems to make more informed decisions. Future developments may focus on integrating VLMs and LLMs with multi-sensor fusion frameworks to enhance their ability to process unstructured, multi-modal data. Additionally, the potential for large-scale pretraining on diverse datasets could improve generalization across various driving scenarios, paving the way for more intelligent and adaptable autonomous systems.

\section{Conclusion}
This paper provides a comprehensive review of multi-sensor fusion techniques for autonomous driving, categorizing methods into data-level, feature-level, object-level, and decision-level fusion. By analyzing state-of-the-art algorithms, we highlight the strengths and limitations of each fusion strategy, emphasizing the importance of feature-level fusion in current applications. The formulation of multi-sensor fusion provides a structured framework for understanding and implementing these methods effectively.

Looking ahead, advancements in hardware, datasets, and algorithmic innovations will continue to drive progress in multi-sensor fusion research. The integration of end-to-end frameworks, combined with the capabilities of VLMs and LLMs, holds significant promise for enhancing the robustness and scalability of autonomous systems. Future work will need to address challenges such as real-time performance, interpretability, and adaptability to diverse driving environments, ensuring the safe and reliable deployment of autonomous vehicles in real-world scenarios.

\bibliographystyle{IEEEtran}

\bibliography{main.bib}

\end{document}